\documentclass[10pt,twocolumn,letterpaper]{article}

\usepackage[dvipsnames,table]{xcolor}
\usepackage[pagenumbers]{cvpr} 
\usepackage{multirow}
\usepackage{graphicx}

\definecolor{cvprblue}{rgb}{0.21,0.49,0.74}
\definecolor{c1}{rgb}{1,0.9,0.9} 
\definecolor{c2}{rgb}{0.9,1,0.9} 
\definecolor{bg}{rgb}{0.9,0.9,0.9} 

\usepackage[pagebackref,breaklinks,colorlinks,allcolors=cvprblue]{hyperref}
\def\paperID{14878} 
\def\confName{CVPR}
\def\confYear{2025}

\title{Hybrid Global-Local Representation with Augmented Spatial Guidance for Zero-Shot Referring Image Segmentation}

\author{Ting Liu$^{1,2}$\thanks{Ting Liu is the corresponding author.} \quad Siyuan Li$^{1}$\\
$^{1}$ASGO, School of Computer Science, Northwestern Polytechnical University\\
$^{2}$Shenzhen Research Institute of Northwestern Polytechnical University\\
{\tt\small liuting@nwpu.edu.cn, \tt\small lisiyuan@mail.nwpu.edu.cn
}
}

\begin{document}
\maketitle
\begin{abstract}
Recent advances in zero-shot referring image segmentation (RIS), driven by models such as the Segment Anything Model (SAM) and CLIP, have made substantial progress in aligning visual and textual information. Despite these successes, the extraction of precise and high-quality mask region representations remains a critical challenge, limiting the full potential of RIS tasks. In this paper, we introduce a training-free, hybrid global-local feature extraction approach that integrates detailed mask-specific features with contextual information from the surrounding area, enhancing mask region representation. To further strengthen alignment between mask regions and referring expressions, we propose a spatial guidance augmentation strategy that improves spatial coherence, which is essential for accurately localizing described areas. By incorporating multiple spatial cues, this approach facilitates more robust and precise referring segmentation. Extensive experiments on standard RIS benchmarks demonstrate that our method significantly outperforms existing zero-shot RIS models, achieving substantial performance gains. We believe our approach advances RIS tasks and establishes a versatile framework for region-text alignment, offering broader implications for cross-modal understanding and interaction. Code is available at \href{https://github.com/fhgyuanshen/HybridGL}{https://github.com/fhgyuanshen/HybridGL}.

\end{abstract}    
\section{Introduction}
\label{sec:intro}
 
Referring image segmentation (RIS) is a critical task in computer vision, where the goal is to segment a specific object or region in an image based on a natural language expression. This task is essential for applications such as visual search, robot perception, and human-computer interaction. Recent advancements, particularly models like the Segment Anything Model (SAM)~\cite{Kirillov_2023_ICCV} and CLIP~\cite{radford2021learning}, have significantly advanced zero-shot RIS, enabling object segmentation without the need for labeled data or task-specific training. In typical RIS pipelines, SAM generates a set of mask proposals, and CLIP extracts visual features for each mask region. For each mask, a similarity score is computed with the referring text, and the mask with the highest score is selected as the final prediction.

\begin{figure}
     \centering
     \includegraphics[width=1\linewidth]{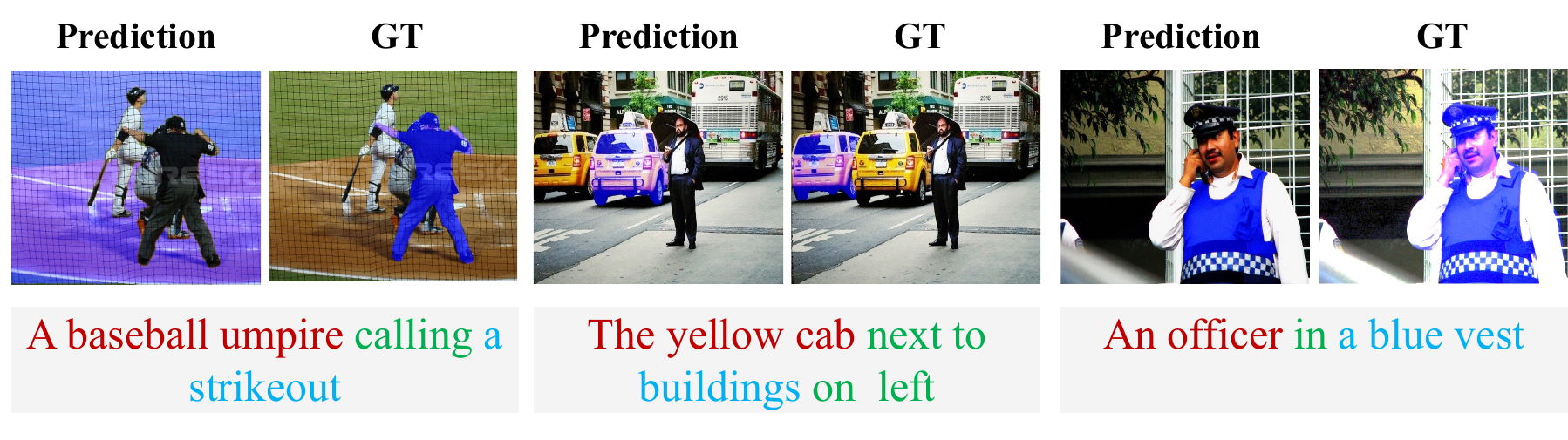}
     \vspace{-4mm}
     \caption{Common issues in existing methods: 1) Inaccurate mask feature extraction; 2) Incorrect spatial localization; 3) Incomplete segmentation.}
     \label{fig:wrongexample}
     \vspace{-6mm}
 \end{figure}
 
Despite the promising performance of existing methods, the accurate extraction of mask representations that align well with referring text remains underexplored. Most existing approaches either mask out areas outside the mask region or crop the image to focus solely on the mask region before feeding it into CLIP, which primarily concentrates on local features while often neglecting the essential surrounding context. Some works~\cite{yu2023zero,suo2023text,ni2023ref} have sought to introduce global surrounding context features. However, such methods remain simplistic and fail to fully capture the complex interplay between the mask and its context, limiting their ability to accurately match referring expressions with the correct mask regions, as shown in \cref{fig:wrongexample}.

In this paper, we propose a novel hybrid global-local feature extraction approach to enhance mask region feature extraction, achieving a more precise and contextually rich mask representation without any additional training. Our approach seamlessly integrates local and global features, capturing both region-specific and context-aware information for each mask. Specifically, we design two complementary branches within CLIP to extract local region-specific visual features and broader context-aware visual features for each mask. 
Features from the global branch are progressively fused into the local branch to generate a hybrid feature representation. This hybrid fusion allows the visual encoder to automatically capture and interact with the complementary information from both branches, yielding a more contextually enriched mask feature.

Another challenge in referring semantic segmentation lies in the use of spatial relationships within referring expressions (e.g., ``left of", ``bottom of") to describe objects, as shown in \cref{fig:wrongexample}. These spatial cues introduce complexity, as they require both the recognition of object locations and the relationship between them. Capturing and aligning these spatial descriptions with visual features is inherently difficult. While referring text offers rich contextual information about spatial relationships, effectively integrating this context with visual data remains a significant challenge. Without a mechanism to explicitly model and align this spatial information, accurately matching the textual description to the correct visual region becomes problematic. Moreover, directly computing the similarity between the extracted visual mask features and text features can lead to ambiguity in the segmentation, as the mask features may capture a mixture of information from multiple objects or regions. it could mistakenly segment only part of the target region, rather than the full object. To alleviate these limitations, we introduce an augmentation approach by introducing several spatial guidance including spatial relationships, coherence, and positional cues.   


Our experiments show that the proposed method significantly outperforms several zero-shot baselines and weakly-supervised referring segmentation methods, achieving significant accuracy improvements. The proposed framework offers a powerful, efficient approach to zero-shot referring image segmentation, with strong potential for practical deployment in real-world scenarios. The contributions of this paper can be summarized as follows:
\begin{itemize} 

\item We propose an innovative hybrid global-local feature extraction approach for RIS, enhancing mask region representation without additional training requirements. 

\item We introduce a spatial guidance augmentation strategy that leverages spatial relationships, coherence, and positional cues to mitigate segmentation ambiguity. 

\item Extensive experiments on the four public datasets, demonstrate that our method significantly outperforms existing state-of-the-art zero-shot semantic segmentation approaches.
\end{itemize}


\begin{figure*}
  \centering
\includegraphics[width=0.9\textwidth]{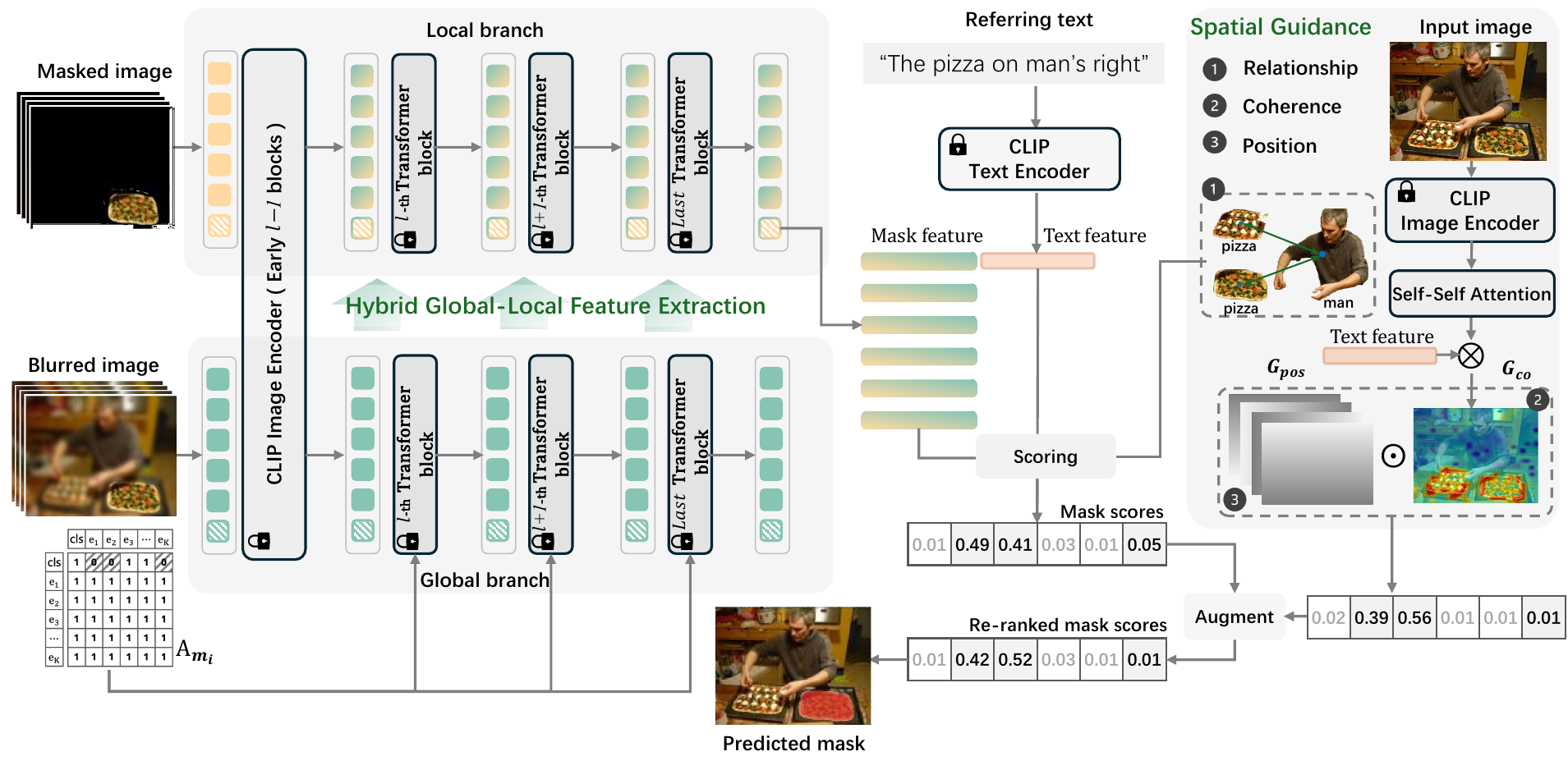}
  \vspace{-5mm}
  \caption{The proposed framework combines hybrid global-local feature extraction with multiple spatial guidance mechanisms to improve zero-shot referring image segmentation, using mask proposals generated by SAM. By leveraging both broad context and local details, and enhancing segmentation with spatial guidance, the framework effectively augments the segmentation of target based on textual descriptions.}
  \vspace{-5mm}
  \label{fig:framework}
\end{figure*}


\section{Related Work}
\label{sec:relatedwork}
{\bf Referring Image Segmentation.} Referring image segmentation is a visual grounding task that requires the model to understand a natural language expression describing a specific object within an image and accurately segment that object from the rest of the scene~\cite{hu2016segmentation}. The goal of this task is to bridge the gap between visual and textual modalities, enabling more sophisticated interactions with visual content. Fully supervised methods~\cite{li2024muti,xu2023meta,li2024smvt,chng2024mask,huang2023referring,shang2024prompt,liu2024rotated,shah2024lqmformer} have achieved impressive performance in this area by effectively integrating information from both images and text descriptions. These methods rely on large datasets with detailed annotations, where each object mentioned in the text is precisely segmented in the corresponding image~\cite{wang2024learning,wang2024unveiling}. Currently, some weakly supervised methods~\cite{strudel2022weakly,lee2023weakly,nag2025safari,Dai_2024_CVPR} can learn and perform segmentation using a smaller amount of labeled data. However, the requirement for such detailed and expensive human-annotated data limits the scalability and applicability of supervised approaches.\\
{\bf Foundation Model in Image Segmentation .} Recent research has shown that segmentation knowledge can emerge from pre-trained foundation models (FMs)~\cite{bommasani2021opportunities} such as CLIP and Stable Diffusion~\cite{rombach2022high}. Though the standard CLIP model excels at recognizing object appearances, it falls short in grasping their exact locations. However, MaskCLIP~\cite{zhou2022extract} demonstrates that it is possible to adapt CLIP for segmentation purposes by making minor adjustments to its attention-pooling mechanism. Other works delve into various strategies for refining attention mechanisms~\cite{li2023clip,bousselham2024grounding,hajimiri2024pay}, enhancing the model's localization capability. Specifically, SAM has shown promising capabilities in object segmentation in many works~\cite{zou2024segment,ravi2024sam,yang2023track,deng2023segment,ma2024segment,xie2023edit,liu2023matcher}. Overall, these findings suggest that FMs can serve as a valuable resource for zero-shot segmentation and related tasks.\\
{\bf Zero-shot Referring Image Segmentation.} To address the limitations of fully supervised methods, zero-shot referring image segmentation has gained significant attention. Zero-shot approaches aim to perform segmentation without requiring any labeled data for the target objects, making them highly flexible and scalable. One notable method in this domain is Global-Local~\cite{yu2023zero}, which leverages pre-trained models like FreeSOLO~\cite{wang2022freesolo} and CLIP to achieve zero-shot segmentation.  Another approach, CaR~\cite{sun2024clip}, further enhances this process by recurrently applying CLIP to refine the segmentation mask iteratively. The introduction of Segment Anything Model (SAM) has marked a significant milestone in zero-shot referring image segmentation. For example, Ref-Diff~\cite{ni2023ref} utilizes diffusion models to better understand the relationship between image and text pairs, leading to more accurate and context-aware segmentations. TAS~\cite{suo2023text} uses a captioner, BLIP2~\cite{li2023blip}, to provide additional context and enhance the segmentation results by generating descriptive captions for the images. Pseudo-RIS~\cite{yu2024pseudo} modifies the Global-Local's pipeline and incorporates a captioner like CoCa~\cite{yu2022coca} for unsupervised training, leveraging the rich knowledge of current foundation models to improve segmentation accuracy. 

\textbf{Note} that although we also adopt both local and global features, our approach differs significantly from Global-Local~\cite{yu2023zero} in terms of both the methods for local and global feature extraction and the fusion strategy of these features. Distinct from previous methods, our work introduces a novel, training-free approach to extracting mask representations using a hybrid feature extraction scheme. Furthermore, we incorporate multiple spatial guidance mechanisms to enhance semantic coherence and spatial alignment, improving the referring segmentation process.

\section{Method}

\subsection{Overview}

Given an input image $ I \in \mathbb{R}^{H \times W \times 3} $, where $ H $ and $ W $ are the height and width of the image, respectively, and a referring text $ T $, which describes the target object or region, the goal is to predict the segmentation mask $ m^* \in \{0, 1\}^{H \times W} $ that corresponds to the region in $ I $ described by $ t $. We first adopt SAM (Segment Anything Model)~\cite{Kirillov_2023_ICCV} to generate a set of mask proposals $ M = \{ m_1, m_2, \dots, m_n \} $, where each $ m_i $ is a binary mask highlighting a potential segment region in $ I $. 

For each mask $ m_i \in M $, we use the CLIP visual encoder to extract our proposed hybrid global-local features  $ \mathbf{x}_i \in \mathbb{R}^d $. To match the referring expression with the visual features, we encode the referring expression $ t $ using the CLIP text encoder $\phi_\text{text}(\cdot)$ to obtain the textual feature $ \mathbf{f}_t \in \mathbb{R}^d $, adopting same strategy with~\cite{yu2023zero}. Subsequently, we can compute the cosine similarity between $ \mathbf{f}_t $ and each feature $ \mathbf{x}_i $, obtaining the semantic alignment score \( \mathbf{S}^s_{m_i}\) for each mask $m_i$. Furthermore, we introduce a spatial guidance augmentation approach. By leveraging several spatial cues, we augment the scoring mask process to prioritize semantically and spatially aligned masks. Our method integrates several spatial cues, specifically spatial coherence guidance, spatial position guidance, and spatial relationships guidance, to enhance the mask scoring process. The overall framework is shown in \cref{fig:framework}.

\subsection{Hybrid Global-Local Feature Extraction}

To extract a more precise mask feature for mask $m_i$, we propose the Hybrid Global-Local Feature Extraction method. This approach captures both region-specific and context-aware features, significantly enhancing mask feature extraction. The method consists of two branches: a local branch for region-specific feature extraction and a global branch for context-aware feature extraction. Features from the global branch are progressively fused into the local branch to generate a hybrid feature representation. This hybrid fusion allows the model to automatically capture the complementary information from the two branches. 

\noindent \textbf{Local Feature Extraction.} For the local region-specific feature extraction, we apply the binary mask $ m $ to the input image $ I $, creating a masked image $ I_\text{local} = I \cdot m_i $. This ensures that only the relevant region contributes to feature extraction. The masked image $ I_\text{local} $ is then processed through the CLIP image encoder to obtain the region-specific features. 
Unlike most existing methods, we do not crop the mask region, ensuring that the input image remains well-aligned with the input of the global branch. This allows the naturally aligned features to be hybridized and deeply fused for enhanced feature extraction.

\noindent \textbf{Global Feature Extraction.} For the context-aware global feature extraction, we use the original image $ I $ and apply a Gaussian blur to the non-mask regions, thereby emphasizing the relevant areas while preserving essential contextual information. To further direct the model's focus onto the mask region, we introduce an attention mask $\mathbf{A}_{m_i} $ during the self-attention operation in the last $l$ layers of the CLIP transformer-based architecture as shown in \cref{fig:framework}. Considering the CLS token is embedded into the shared visual-textual space as the visual feature, we nullify the attention scores between the CLS token and image tokens outside the masked region, effectively preventing non-relevant areas from contributing to the attention on the CLS token. 

\noindent \textbf{Hybrid Feature Extraction.} To enhance the fusion of local and global features, we propose a hybrid fusion approach that enables the attention mechanism in the network to effectively and automatically capture the complementary information between these two features. 

Given that local features primarily focus on the mask region, which typically contains the most relevant information for the task, the global branch is intended to complement the local branch by providing a broader context. While the global branch captures a wider spatial range and provides essential contextual information beyond the mask, it may also introduce irrelevant information from regions outside the target. To maximize the benefits of both local and global features, it is crucial to selectively integrate them, ensuring that the global context enhances the local representation without diluting its focus on the target region.

To achieve this, we apply a token mask to the global features during the fusion. Specifically, we mask out image tokens outside the mask region to ensure that only the relevant tokens contribute to the feature extraction process. The fusion of local and global features at layer $ l $ is performed as follows:   
\begin{equation}
    \mathbf{x}_\text{hybrid}^{(l)} = \mathbf{x}_\text{local}^{(l-1)} + \beta \cdot \left( \mathbf{x}_\text{global}^{(l-1)} \cdot \mathbf{B}_{m_i} \right),
\end{equation} where $ \mathbf{x}_\text{local}^{(l-1)} $ and $\mathbf{x}_\text{global}^{(l-1)}$ represent the local and global features derived from the ${l-1}$-th layer, respectively. The hyper-parameter $\beta$ is the relative contribution of the global feature to the fusion process. To ensure that only relevant tokens contribute to the feature fusion, we apply a mask $\mathbf{B}_{m_i} \in \{0,1\}^K$ on the global feature, where each feature contains $K$ image token features, effectively excluding tokens corresponding to regions outside the mask. The resulting hybrid feature $ \mathbf{x}_\text{hybrid}^{(l-1)} $ is then passed as input to the $l$-th layer of the CLIP image encoder, denoted $\phi^l_{\text{visual}}$, to produce a feature representation that integrates both global and local information: \begin{equation}
\mathbf{x}_\text{local}^{(l)} = \phi^l_{\text{visual}}(\mathbf{x}_\text{hybrid}^{(l)}) .
\end{equation} 

The local feature $\mathbf{x}_\text{local}^{(L)} $ obtained from the last encoder layer $ L $, which effectively combines localized and global contextual information, serves as the final hybrid feature for the mask region. 
   
\noindent \textbf{Semantic Alignment Score.} We then extract hybrid visual features \(\mathbf{X}_{\text{hybrid}} = \{\mathbf{x}_1, \mathbf{x}_2, \dots, \mathbf{x}_n\}\) for each mask, where each \(\mathbf{x}_i\) corresponds to a mask proposal \(m_i\). After extracting a referring text feature \(\mathbf{f}_t\) from the CLIP text encoder, we calculate the cosine semantic alignment score between the referring text and each \(\mathbf{x}_i\): 
\begin{equation}
    \mathbf{S}^s_{m_i} = \text{cos}( \phi_\text{text}(t), x_i).
    \label{eq:semantic_score}
\end{equation}
  
\subsection{Spatial Guidance Augmentation} We combine multiple spatial guidance mechanisms, including relationships, coherence, and position, which are detailed in the following.

\noindent \textbf{Spatial Relationship Guidance.}  To utilize the spatial relationships described in the referring text, we first parse the text into objects and spatial relations, following a method similar to~\cite{subramanian2022reclip}. For instance, objects might include phrases like ``the pizza" or ``man" while spatial relations describe the positioning between objects, such as ``right". 

With the extracted hybrid mask features $x_i$ for each mask $m_i$, we compute the semantic alignment score $\mathcal{P}(p, m_i) = \text{cos}( \phi_\text{text}(p), x_i)$, which measures how well the mask satisfies a given object $p$ (e.g., ``the pizza"). We then select the top $k$ masks based on this score and apply the softmax function to normalize these selected scores, ensuring that they sum to one. This strategy prioritizes the most relevant and semantically aligned mask proposals while also enhancing computational efficiency for subsequent operations.  

To model the spatial relationships, a spatial relation function $\mathcal{R}(p,q)$ quantified the relationship between two parsed objects $p$ and $q$ is defined as:
\begin{equation}
    \mathcal{R}((p,m_i) ,(q,m_j)) = \begin{cases} 
1 & \text{if } p \text{ satisfies the relation with } q, \\
0 & \text{otherwise} ,
\end{cases}
\end{equation}
where the spatial relation between $p$ and $q$ can be any of the defined relations like ``left", ``right", ``top", ``bottom", ``within", ``smaller", or ``bigger". The satisfaction of the spatial relation is computed based on the position and size of the corresponding masks $m_i$ and $m_j$ for $p$ and $q$.

Finally, we combine the probability of each object with the spatial relation probabilities to identify the target object. The overall likelihood of mask $m_i$ being the target $p$ is given by:
\begin{equation}
     \mathbf{S}^s_{m_i} = \sum_{m_j} \mathcal{P}(p, m_i) \cdot \mathcal{R}((p,m_i) ,(q,m_j)) \cdot \mathcal{P}(q, m_j),
\end{equation}
where $\mathbf{S}^s_{m_i}$ is the final probability of $m_i$ being the correct target, combining both its predicate satisfaction and its spatial relationships with other objects. If the referring text contains those spatial relations, the score of each mask is computed accordingly. Otherwise, $\mathbf{S}^s_{m_i}$ is directly computed using Equation.~\ref{eq:semantic_score}.

\noindent \textbf{Spatial Coherence Guidance.}  To enhance spatial coherence, we generate a spatial localization guidance \( \mathbf{G}_{\text{co}} \in [0,1]^{H \times W} \) by using the referring text feature \( (t) \) to identify the target region.  Specifically, we apply the algorithm proposed in \cite{bousselham2024grounding}, which leverages self-attention mechanisms for expression localization and segmentation. This guidance map is constructed by calculating the similarity between \( \phi_\text{text}(t) \) and each visual token embedding 
derived from the CLIP image encoder with the self-attention mechanisms, yielding a localization map that broadly highlights areas corresponding to the target object or region. 
Consequently, positions with values closer to 1 represent higher similarity, thereby indicating regions more likely aligned with the target described by the text feature.

\noindent \textbf{Spatial Position Guidance.} To incorporate spatial positions that align with the referring expression context, similar with~\cite{ni2023ref}, we introduce a set of positional guidance matrices \( \mathbf{G}_{\text{pos}} \in [0, 1]^{H \times W} \), each representing a position within the image. Here, \( pos \in \{ \textit{top}, \textit{bottom}, \textit{left}, \textit{right}, \textit{middle} \} \) denotes specific spatial attributes, such as ``top" for upper regions. $\mathbf{G}_{\text{pos}}$ is computed by:
\begin{equation}
    \mathbf{G}_{\text{pos}} = 
    \begin{cases} 
    \mathcal{E}(\text{pos}) & \text{if position } $pos$ \text{ appears in } t, \\
    \mathbf{1}_{H \times W} & \text{if no position $pos$ appears in } t,
    \end{cases} 
\end{equation}
where \( \mathcal{E}(\text{pos}) \) represents the emphasis function that returns a value between 0 and 1 based on the position's emphasis. To generate these matrices, we compute the distance of each pixel to the corresponding region defined by \( \text{pos} \), normalize the distances to the range \( [0, 1] \), and directly assign these values to the positional guidance matrix. This structure reflects a gradual transition in spatial emphasis across the image, enabling the model to focus on regions by the referring expression's context.
 
In this manner, we define \( \mathbf{G}_{\text{pos}} \) for each desired direction. Once the $pos$ appears in the referring text, we incorporate this spatial guidance into our spatial coherence guidance \( \mathbf{G}_{\text{co}} \) by performing an element-wise multiplication. For each direction \( \text{dir} \), the spatial guidance matrix \( \mathbf{G}\) is defined as:
\begin{equation} 
\mathbf{G} = \mathbf{G}_{\text{co}} \odot \mathbf{G}_{\text{pos}}.
\end{equation}
   

\noindent \textbf{Spatial Guidance Score. } The spatial guidance score \( \mathbf{S}^g_{m_i} \) for each mask proposal \( m_i \) is computed as the difference between the mean spatial guidance values within the mask. Specifically, we define the spatial guidance score as:
\begin{equation}
\mathbf{S}^g_{m_i} = \frac{\text{Sum}(\mathbf{G} \odot m_i)}{\text{Sum}(m_i)} - \lambda \cdot  \frac{\text{Sum}(\mathbf{G} \odot (1 - m_i))}{\text{Sum}(1 - m_i)},
\end{equation}
where $\lambda$ is a hyperparameter that controls the importance of the negative score. This score \( \mathbf{S}^g_{m_i} \) serves as a spatial coherence measure, with larger values indicating better spatial alignment between the mask proposal and the referring expression. 

Finally, we apply softmax normalization to the semantic alignment score \( \mathbf{S}^s_{m_i} \) and spatial guidance score \( \mathbf{S}^g_{m_i} \) for all masks, and fuse them together.
The final score for each mask \( m_i \) is obtained by: 
\begin{equation} 
\mathbf{S}_{m_i} = (1-\alpha) \mathbf{S}^s_{m_i} + \alpha \mathbf{S}^g_{m_i},
\end{equation}
where $\alpha$ controls the trade-off between semantic alignment and spatial guidance. The final referring semantic segmentation result \( m^* \)  is generated by selecting the mask with the highest mask score. 

This comprehensive approach allows us to prioritize mask proposals that are not only semantically aligned with the text feature but also spatially consistent with the specified direction. Thus, masks are refined based on both semantic and spatial coherence, enhancing the accuracy of the selected mask with respect to the referring expression’s spatial context. 

\section{Experiments}

\begin{table*}[h]
\centering
\resizebox{\textwidth}{!}{%
\begin{tabular}{c|ccc|ccc|ccc|cc}
\toprule
\multirow{2}{*}{Metric} & {\multirow{2}{*}{Method}} & {\multirow{2}{*}{Vision Backbone}} & \multirow{2}{*}{Pre-trained Model} & \multicolumn{3}{c|}{RefCOCO}      & \multicolumn{3}{c|}{RefCOCO+}     & \multicolumn{2}{c}{RefCOCOg} \\
    &                  &                          &                             & val       & testA     & testB     & val       & testA     & testB     & val           & test         \\ 
\hline
\multirow{12}{*}{oIoU}  & \multicolumn{3}{l|}{\textit{zero-shot methods w/ additional training}}       &  &  &  &  &  &  &  &     \\
    & {Pseudo-RIS}~\cite{yu2024pseudo}     & {ViT-B
    }             & SAM, CoCa, CLIP              & 37.33     & 43.43     & 31.90     & 40.19     & 46.43     & 33.63     & 41.63         & 43.52        \\
    & {VLM-VG}~\cite{wang2024learning}     & {R101}   & ${\text{COCO}}^{*}$, ${\text{VLM-VG}}^{*}$ & 45.40     & 48.00     & 41.40     & 37.00     & 40.70     & 30.50     & 42.80         & 44.10        \\ 
    \cline{2-12} 
    & \multicolumn{3}{l|}{\textit{zero-shot methods w/o additional training}}                          &  &  &  &  &  &  &  &     \\
    & {Grad-CAM}~\cite{2017gradcam}       & {R50}                    & SAM, CLIP             & 23.44     & 23.91     & 21.60     & 26.67     & 27.20     & 24.84     & 23.00         & 23.91        \\
    & {MaskCLIP}~\cite{zhou2022extract}      & {R50}                 & SAM, CLIP             & 20.18     & 20.52     & 21.30     & 22.06     & 22.43     & 24.61     & 23.05         & 23.41        \\
    & {Global-Local}~\cite{yu2023zero}   & {R50} & FreeSOLO, CLIP             & 24.58     & 23.38     & 24.35     & 25.87     & 24.61     & 25.61     & 30.07         & 29.83        \\
    & {Global-Local}~\cite{yu2023zero}   & {R50}                  & SAM, CLIP                  & 24.55     & 26.00     & 21.03     & 26.62     & 29.99     & 22.23     & 28.92         & 30.48        \\
    & {Global-Local}~\cite{yu2023zero}   & {ViT-B
    }              & SAM, CLIP             & 21.71     & 24.48     & 20.51     & 23.70     & 28.12     & 21.86     & 26.57         & 28.21        \\
    \rowcolors{2}{gray!20}{white}
    & \cellcolor{bg}{Ref-Diff}~\cite{ni2023ref}       & \cellcolor{bg}{ViT-B
    }           & \cellcolor{bg}SAM, SD, CLIP              & \cellcolor{c2}\underline{35.16}     & \cellcolor{c2}\underline{37.44}     & \cellcolor{c2}\underline{34.50}     & \cellcolor{c2}\underline{35.56}     & 38.66     & \cellcolor{c1}\textbf{31.40}     & \cellcolor{c2}\underline{38.62}         & \cellcolor{c2}\underline{37.50}        \\
    & {TAS}~\cite{suo2023text}            & {ViT-B
    }              & SAM, BLIP2, CLIP           & 29.53     & 30.26     & 28.24     & 33.21     & \cellcolor{c2}\underline{38.77}     & 28.01     & 35.84         & 36.16        \\
    & \cellcolor{bg}{Ours}           & \cellcolor{bg}{ViT-B
    }               & \cellcolor{bg}SAM,CLIP                    & \cellcolor{c1}\textbf{41.81} & \cellcolor{c1}\textbf{44.52} & \cellcolor{c1}\textbf{38.5} & \cellcolor{c1}\textbf{35.74} & \cellcolor{c1}\textbf{41.43} & \cellcolor{c2}\underline{30.9} & \cellcolor{c1}\textbf{42.47} & \cellcolor{c1}\textbf{42.97}    \\ 
\hline \hline
\multirow{16}{*}{mIoU}  & \multicolumn{3}{l|}{\textit{weakly-supervised methods}}       &  &  &  &  &  &  &  &     \\
    & {CLRL}~\cite{lee2023weakly}     & {ViT-B
    }             & -              & 31.06 & 32.30 & 30.11 & 31.28 & 32.11 & 30.13 & 32.88 &  -       \\
    & {PPT}~\cite{Dai_2024_CVPR}     & {ViT-B}   & SAM & 46.76 & 45.33 & 46.28 & 45.34 & 45.84 & 44.77 & 42.97 & -       \\
    \cline{2-12} 
    & \multicolumn{3}{l|}{\textit{zero-shot methods w/ additional training}}       &  &  &  &  &  &  &  &     \\
    & {Pseudo-RIS}~\cite{yu2024pseudo}     & {ViT-B
    }             & SAM, CoCa, CLIP              & 41.05     & 48.19     & 33.48     & 44.33     & 51.42     & 35.08     & 45.99         & 46.67        \\
    & {VLM-VG}~\cite{wang2024learning}     & {R101}   & ${\text{COCO}}^{*}$, ${\text{VLM-VG}}^{*}$ & 49.90     & 53.10     & 46.70     & 42.70     & 47.30     & 36.20     & 48.00         & 48.50        \\
    \cline{2-12} 
    & \multicolumn{3}{l|}{\textit{zero-shot methods w/o additional training}}                          &  &  &  &  &  &  &  &     \\
    & {Grad-CAM}~\cite{2017gradcam}       & {R50}              & SAM, CLIP             & 30.22     & 31.90     & 27.17     & 33.96     & 25.66     & 32.29     & 33.05         & 32.50        \\
    & {MaskCLIP}~\cite{zhou2022extract}      & {R50}           & SAM, CLIP             & 25.62     & 26.66     & 25.17     & 27.49     & 28.49     & 30.47     & 30.13         & 30.15        \\
    & {Global-Local}~\cite{yu2023zero}   & {R50}               & FreeSOLO, CLIP        & 26.70     & 24.99     & 26.48     & 28.22     & 26.54     & 27.86     & 33.02         & 33.12        \\
    & {Global-Local}~\cite{yu2023zero}   & {R50}                  & SAM, CLIP                   & 31.83     & 32.93     & 28.64     & 34.97     & 37.11     & 30.61     & 40.66         & 40.94        \\
    & {Global-Local}~\cite{yu2023zero}   & {ViT-B
    }              & SAM, CLIP             & 33.12     & 36.52     & 29.58     & 35.29     & 39.58     & 31.89     & 40.08         & 40.74        \\
    & {CaR}~\cite{sun2024clip}            & {ViT-B  and ViT-L}
    & CLIP                        & 33.57     & 35.36     & 30.51     & 34.22     & 36.03     & 31.02     & 36.67         & 36.57        \\
    & {Ref-Diff}~\cite{ni2023ref}       & {ViT-B
    }           & SAM, SD, CLIP               & 37.21     & 38.40     & 37.19     & 37.29     & 40.51     & 33.01     & 44.02         & 44.51        \\
    & \cellcolor{bg} {TAS}~\cite{suo2023text}            & \cellcolor{bg} {ViT-B
    }              & \cellcolor{bg} SAM, BLIP2, CLIP            & \cellcolor{c2}\underline{39.84}     & \cellcolor{c2}\underline{41.08}     & \cellcolor{c2}\underline{36.24}     & \cellcolor{c1}\textbf{43.63}     & \cellcolor{c1}\textbf{49.13}     & \cellcolor{c2}\underline{36.54}     & \cellcolor{c2}\underline{46.62}         & \cellcolor{c2}\underline{46.80}        \\
    & \cellcolor{bg} Ours  & \cellcolor{bg} ViT-B
                 & \cellcolor{bg} SAM, CLIP                    & \cellcolor{c1}\textbf{49.48} & \cellcolor{c1}\textbf{53.37} & \cellcolor{c1}\cellcolor{c1}\textbf{45.19} & \cellcolor{c2}\underline{43.40} & \cellcolor{c1}\textbf{49.13} & \cellcolor{c1}\textbf{37.17} & \cellcolor{c1}\textbf{51.25}     & \cellcolor{c1} \textbf{51.59}    \\ 
\bottomrule
\end{tabular}%
}
\vspace{-2mm}
\caption{Comparisons with the SOTA zero-shot approaches on RefCOCO, RefCOCO+, and RefCOCOg datasets. The best two results under the same setting, w/o additional training, are highlighted in \colorbox{c1}{\textbf{bold}} and \colorbox{c2}{\underline{underlined}}, respectively. * indicates the extra dataset used to train the model.} 
    \vspace{-5mm}
\label{tab:acc-cmp}
\end{table*}

\subsection{Datasets and Metrics}
To evaluate the proposed method, we use several main RIS datasets: RefCOCO~\cite{nagaraja2016modeling}, RefCOCO+~\cite{nagaraja2016modeling}, and RefCOCOg~\cite{mao2016generation}. These datasets are derived from the MSCOCO~\cite{lin2014microsoft} dataset and feature images annotated with detailed referring expressions that pinpoint specific objects or regions. Each dataset brings unique aspects to the referring expressions it contains. RefCOCO often includes positional information like ``left" and ``right," which is banned in RefCOCO+, and RefCOCOg has more elaborate sentence structures. We also evaluate it on PhraseCut~\cite{wu2020phrasecut} dataset, which introduces structured textual descriptions that detail attributes, categories, and relationships among objects.
We employ two primary metrics to evaluate the performance: overall Intersection over Union (oIoU) and mean Intersection over Union (mIoU). oIoU assesses the total overlap between predicted and ground truth regions relative to their combined area, making it particularly stringent for inaccuracies in larger segments. On the other hand, mIoU calculates the average overlap for each individual instance, ensuring a balanced consideration of performance across objects of varying sizes. Together, these metrics provide a robust framework for assessing the effectiveness of the proposed method in RIS tasks.
\subsection{Implementation Details}
The experiments were run on a single NVIDIA RTX 3090 GPU. Following previous work~\cite{suo2023text,ni2023ref}, we use the default ViT-H SAM model, and the hyperparameters ``predicted iou threshold" and ``stability score threshold" were set to 0.7, the ``points per side" was set to 8. We use CLIP with ViT-B/16 backbone in both hybrid feature extraction and spatial coherence guidance. We set $\alpha$ and $\beta$  to 0.6 and 2, respectively, for all datasets. The hyperparameter $\lambda$ is empirically set to 9 to balance the propensity of both excessively large and small mask regions receiving disproportionately high scores. When the referring text includes ``big" or ``small",  $\lambda$ is adjusted to 3 or 14, respectively.

\subsection{Results}
In \cref{tab:acc-cmp}, we evaluate our model on RefCOCO, RefCOCO+, and RefCOCOg and compare it with other state-of-the-art (SOTA) zero-shot models. Here, we evaluate the MaskCLIP, Grad-CAM methods by computing the similarity between their feature maps and SAM's mask proposals. Since the original Global-Local method used FreeSOLO as the mask extractor, we have re-evaluated the performance of Global-Local using SAM as the mask extractor, across different backbones. We also provide results of Ref-Diff, and TAS using different backbones. Additionally, we include weakly-supervised methods and zero-shot methods with extra training or additional datasets. Our method achieves excellent results on all three datasets. In terms of oIoU, our method improves by 4\%-7\% over SOTA methods on RefCOCO and RefCOCOg, and although it does not achieve the best performance on the testB set of RefCOCO+, it is only 0.5\% lower than the SOTA method. For mIoU, our method outperforms SOTA methods by 4\%-10\% on RefCOCO and RefCOCOg, and it also achieves comparable performance on RefCOCO+. Importantly, our method achieves comparable or even higher mIoU on all three datasets compared to methods that use additional training. This indicates that our method has a more balanced consideration across varying objects. In addition, we present the oIoU and mIoU results on the PhraseCut dataset’s test set in \cref{tab:acc-cmp-pc}. Our method demonstrates superior performance compared to previous methods, achieving the highest average score. 

\begin{table}[h]
    \centering
    \setlength{\tabcolsep}{4mm}\begin{tabular}{c|cc|c}
        \toprule
        Methods         & oIoU      & mIoU      & avg.\\
        \midrule
        Global-Local\cite{yu2023zero}   & 23.64             & -             & 23.64\\
        TAS\cite{suo2023text}           & 25.64             & 24.66         & 25.15\\
        Ref-Diff\cite{ni2023ref}        & 29.42             & \textbf{41.75} & 35.59\\
        Ours                            & \textbf{38.39}    & 36.98         & \textbf{37.69}\\
        \bottomrule
    \end{tabular}
    \caption{Comparison with existing methods on the PhraseCut dataset. } 
    \vspace{-5mm}
    \label{tab:acc-cmp-pc}
\end{table}

\begin{figure*}[h]
  \centering
    \includegraphics[width=0.95\textwidth]{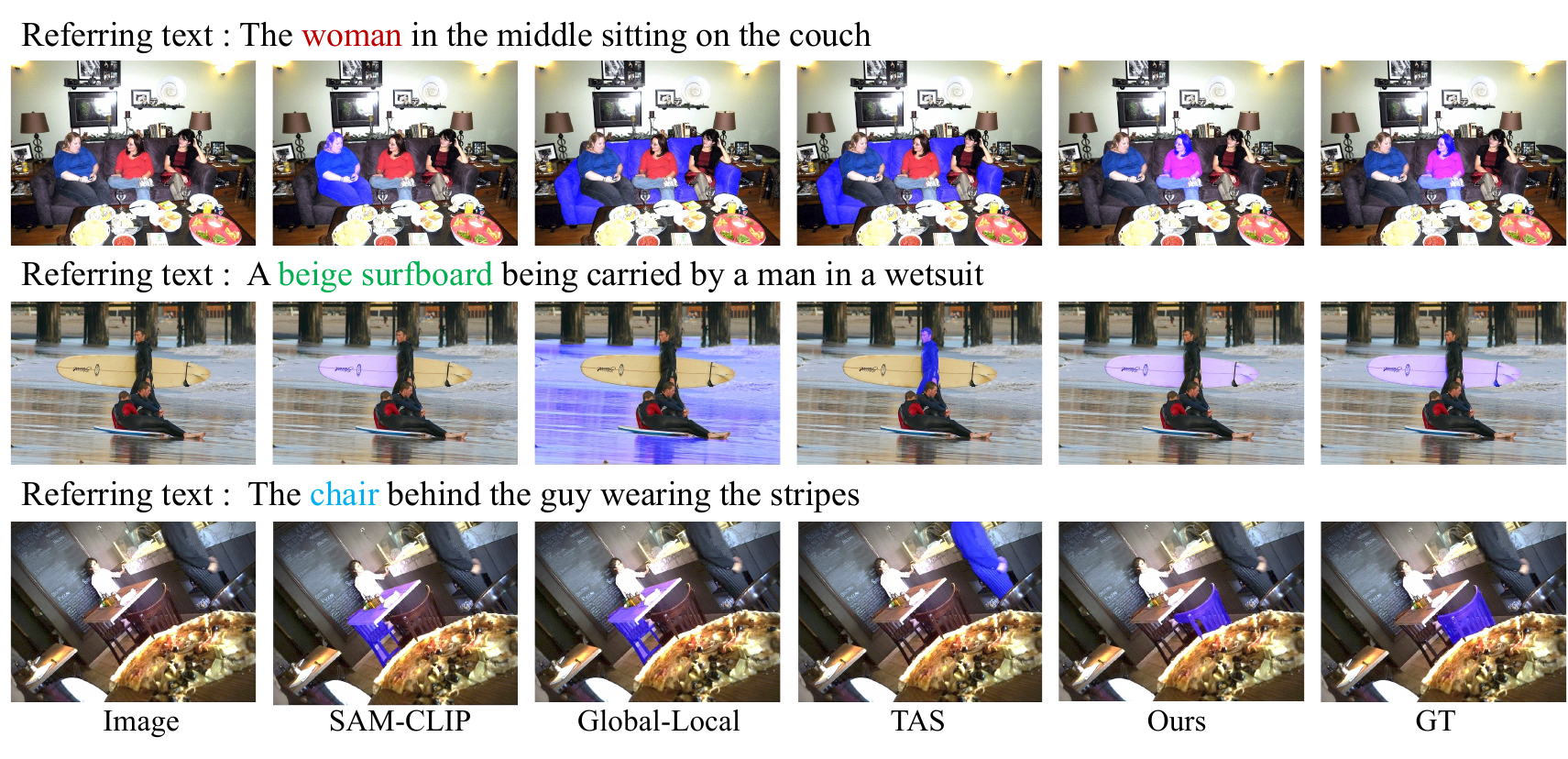}
    \vspace{-5mm}
  \caption{Visual comparisons with existing methods. Our approach achieves more accurate localization and a complete segmentation of the target object.}
  \label{fig:vis}
    \vspace{-5mm}
\end{figure*} 

\textbf{Notably}, our approach relies solely on SAM and CLIP, while TAS additionally incorporates the large BLIP2 model, and Ref-Diff leverages the Stable-Diffusion model; yet, our method still outperforms these more complex models. Besides, TAS and Ref-Diff show better performance in either mIoU or oIoU, whereas our method consistently outperforms both metrics. We present visual comparisons with existing methods in \cref{fig:vis}, highlighting the improvements in segmentation accuracy and the ability to capture spatial relationships.

\subsection{Ablation Study}

To evaluate the effectiveness of different strategies, we conduct extensive ablation studies on the \textit{val} dataset of RefCOCO, RefCOCO+, and RefCOCOg datasets. We compare a range of feature extraction methods and spatial guidance strategies, examining their impact on performance.

\subsubsection{Ablation on Hybrid Feature Extraction}

\noindent\textbf{Local and Global Features.}
We first evaluate the performance of local and global features, including different strategies for global feature extraction. These strategies involve blurring images to reduce the influence of irrelevant regions on the global representation, applying a token mask to exclude non-mask regions, and using an attention mask to minimize the impact of non-mask regions on the CLS token. All these different mask strategies are applied to the last few layers of the CLIP image encoder. As shown in \cref{tab:abla_HF}, the local features (L) consistently outperform the global methods across all datasets in terms of both oIoU and mIoU. Notably, combining a blur operation with the attention mask (G(blur+att\_mask)) results in a significant improvement, approaching the performance of local features on the RefCOCO+ and RefCOCOg datasets. Hence, we adopt this approach to extract global features. 

\begin{table}[h]
    \centering
    \setlength{\tabcolsep}{0.7mm}{
    \begin{tabular}{c|cc|cc|cc}
        \toprule
        \multirow{2}{*}{Method} & \multicolumn{2}{c|}{RefCOCO}      & \multicolumn{2}{c|}{RefCOCO+}     & \multicolumn{2}{c}{RefCOCOg} \\
                  & oIoU  & mIoU & oIoU  & mIoU  & oIoU  & mIoU \\
        \midrule
        L            & 24.65 & 32.64 & 28.16 & 36.44 & 33.63 & 42.35\\ 
        G(blur)      & 16.17 & 22.01 & 18.61 & 25.14 & 22.00 & 31.24\\
        G(tok\_mask) & 20.55 & 31.71 & 21.84 & 33.18 & 19.82 & 32.55\\
        G(att\_mask) & 20.05 & 30.07 & 21.44 & 31.51 & 24.11 & 36.24\\
        G(blur+att\_mask) & 27.63 & 37.12 & 29.75 & 39.94 & 33.03 & 44.45\\ 
        \hline
         G + L ~\cite{yu2023zero}      & 21.71 & 33.12 & 23.70 & 35.29  & 26.57 & 40.08 \\
        G + L       & 28.90 & 37.64 & 32.27 & 41.12 & 38.08 & 47.51\\
        L2G         & 29.93 & 39.17 & 33.17 & 42.55 & 37.71 & 47.20\\
        G2L         & \textbf{31.71} & \textbf{40.27} & 34.44 & 43.40 & \textbf{39.12} & \textbf{48.64}\\
        G2L + L2G   & 31.32    & 40.23   & \textbf{34.63}    & \textbf{43.73}    & 38.43 & 48.22\\ 
        \bottomrule
    \end{tabular}
    }
    \vspace{-0.2cm}
    \caption{Results of different mask feature extraction methods on the \textit{val} split of three datasets. } 
    \vspace{-0.65cm}
    \label{tab:abla_HF}
\end{table}

\noindent\textbf{Hybrid Global-Local Features.}
The intuitive way to fuse the global and local features is to compute a weighted sum of the two. In \cref{tab:abla_HF}, we present experimental results of this fusion strategy, employing the local and global feature extraction method from ``G + L\cite{yu2023zero}", which denotes the weighted sum of G(tok\_mask) and cropped L, while ``G + L” refers to the weighted sum of G(blur+att\_mask) with L. We report both the results with the optimal weight that yielded the highest performance. Our results demonstrate that, due to the effectiveness of our global feature extraction strategy in generating a more accurate global representation, this weighted sum fusion significantly outperforms the previous method. 

To fully exploit the attention mechanism in the model and facilitate a comprehensive interaction between the global and local features, we propose generating a hybrid feature by fusing features from the later layers of one branch's image encoder into the other branch. 
In \cref{tab:abla_HF}, we present the results of different hybrid fusion strategies. The ``L2G" strategy, where local branch features are incorporated into the global branch, shows a significant improvement over ``G + L". This highlights the effectiveness of our hybrid feature extraction method. On the other hand, the ``G2L" strategy, where global features are fused into the local branch, provides the best results overall. This indicates that local features, which are crucial for fine-grained object identification, benefit most from the additional global context, allowing the model to achieve more accurate and contextually informed segmentations. The key difference between ``G2L" and ``L2G" is that the global branch applies an attention mask, while the local branch does not. Global features require a token mask for integration into the local branch, whereas local features can be directly fused into the global branch, resulting in distinct ``G2L" and ``L2G" outputs. We also explore the fusion of both ``L2G" and ``G2L" strategies using a weighted sum operation, but no further improvements are observed. Consequently, we adopt the ``G2L" strategy for hybrid feature extraction in our implementation.

\noindent\textbf{Impact of Starting Layer for Hybrid Fusion} We further investigate the effect of starting the hybrid fusion at different layers of the image encoder on the RefCOCOg dataset. In \cref{tab:abla_masklayer}, we present the results of applying the fusion at various layers to determine which starting layer yields the best performance. Our findings show that initiating fusion at the 9\textit{-th} layer yields the best performance, suggesting that representations at this layer provide the most meaningful and contextually relevant information for effective interaction between global and local features. 
   
\subsubsection{Ablation on Spatial Guidance}
Building upon the proposed hybrid global-local feature extraction method, we further conduct ablation studies to assess the impact of different components in the spatial guidance augmentation approach. Our experiments primarily focus on two datasets: RefCOCO and RefCOCOg.


\begin{table}
    \centering
    \setlength{\tabcolsep}{2.8mm}{
    \begin{tabular}{c|cc|cc}
        \toprule
        \multirow{2}{*}{Start Layer} & \multicolumn{2}{c|}{G2L}     & \multicolumn{2}{c}{L2G} \\
                  & oIoU  &  mIoU  & oIoU  & mIoU \\
        \midrule
        7  &  22.61 & 34.78 & 29.12 & 39.33\\
        8  &  38.16 & 47.82 & 34.67 & 43.62\\
        9  &  \textbf{39.12} & \textbf{48.64} & \textbf{37.61} & \textbf{47.17}\\
        10 &  36.50 & 46.47 & 37.38 & 46.81\\
        11 & 34.12 & 42.26 & 33.28 & 44.06\\
        \bottomrule
    \end{tabular}
    }
    \caption{Results of the starting layer $l$ for hybrid fusion of global and local branches on the \textit{val} split of RefCOCOg dataset.}
    \vspace{-5mm}
    \label{tab:abla_masklayer}
\end{table}

\noindent\textbf{Impact of Spatial Relationship Guidance}
 From ~\cref{tab:gemornot}, we can see that incorporating spatial relationship guidance (Rel) leads to consistent performance improvements across all datasets. However, the improvements are more limited on RefCOCOg, where referring expressions typically contain fewer spatial relationships. The most significant improvement is seen on RefCOCO, where both mIoU and oIoU increase by approximately 4\%. This is attributed to the richer spatial relationship descriptions in the referring expressions, which are better captured by the introduced guidance. Overall, these results highlight the effectiveness of spatial relationship guidance in enhancing referring expression comprehension, especially in datasets with complex and detailed spatial descriptions.

\noindent\textbf{Impact of Spatial Coherence \& Position Guidance}
Spatial coherence guidance \( \mathbf{G}_{\text{co}} \) is introduced to enhance segmentation coherence and mitigate the issue of partial masks being selected for the target object. As a result, improvements in oIoU and mIoU are modest, as only a small subset of data face this issue, which may not be fully reflected in the overall scores.  However, when combined with spatial position guidance \( \mathbf{G}_{\text{pos}} \) together, as shown in \cref{tab:gemornot}, we observe a significant improvement in spatial position awareness, leading to more accurate target localization and enhanced segmentation performance. In referring expressions, spatial positional cues such as ``the left [object]" often specify the target object's position without additional information. By introducing \( \mathbf{G}_{\text{pos}}\), which explicitly encodes spatial positioning information, we improve the alignment between the referring text and the mask prediction.

\begin{table}
    \centering
    \setlength{\tabcolsep}{1.5mm}{
    \begin{tabular}{cccc|cc|cc}
        \toprule
        \multirow{2}{*}{G2L} &\multirow{2}{*}{Rel }  & \multirow{2}{*}{\(\mathbf{G}_{\text{co}}\)} & \multirow{2}{*}{\(\mathbf{G}_{\text{pos}}\) } & \multicolumn{2}{c|}{RefCOCO}    &  \multicolumn{2}{c}{RefCOCOg} \\
                & & & & oIoU  & mIoU & oIoU  & mIoU \\
        \midrule
            \checkmark  &        &            &             & 31.71 & 40.27  & 39.12 & 48.64\\
        \checkmark  &\checkmark  &            &             & 35.68 & 44.29 & 39.68 & 48.99 \\
         \checkmark  & \checkmark  & \checkmark &  & 35.68 & 44.29 & 39.73 & 49.02 \\
        \checkmark  &\checkmark  & \checkmark & \checkmark  & {\textbf{41.81}} & \textbf{49.48} &  {\textbf{42.47}} & {\textbf{51.25}}\\
        
        \bottomrule
    \end{tabular}
    }
    \caption{Ablation study on the different spatial guidance.}
    \vspace{-5mm}
    \label{tab:gemornot}
\end{table}

\section{Conclusion} 

This paper presents a novel, training-free approach to zero-shot referring image segmentation (RIS), addressing challenges in aligning visual masks with referring expressions. Leveraging CLIP and SAM, our hybrid global-local feature extraction method combines mask-specific detail with contextual information to improve mask representation. Further, a spatial guidance augmentation strategy enhances spatial coherence and reduces ambiguities, effectively aligning masks with referring text. Experiments on RefCOCO, RefCOCO+, and RefCOCOg demonstrate that our method achieves substantial gains over zero-shot RIS models.

\noindent \textbf{Acknowledgements.} This work was supported in part by the National Natural Science Foundation of China (No.62106201, No.62376217), and in part by the Guangdong Basic and Applied Basic Research Foundation (No.2025A1515011501). The authors thank the anonymous reviewers for their helpful feedback.
{
    \small
    \bibliographystyle{ieeenat_fullname}
    \bibliography{main}
}


\end{document}